\documentclass[12pt]{article}
\usepackage{amsmath,amsfonts}
\usepackage{graphicx,psfrag,epsf}
\usepackage{enumerate}
\usepackage{natbib}
\usepackage{url} 
\usepackage{booktabs}
\usepackage{tabularx}

\newcommand{\blind}{0}

\def\spacingset#1{\renewcommand{\baselinestretch}%
{#1}\small\normalsize} \spacingset{1}

\usepackage{lscape} 

\usepackage[]{algorithm2e}
\usepackage{bm}
\usepackage{subfigure}
\usepackage{authblk}
\usepackage[format=hang,labelfont=bf]{caption}
\usepackage{url}
\usepackage{hyperref}
\usepackage{todonotes}
\usepackage[capposition=bottom]{floatrow}
\usepackage{paralist} 
\usepackage{xcolor}
\usepackage[toc,page]{appendix}

\usepackage{scalefnt}
\usepackage{bm}
\usepackage{comment}
\newcommand{\ok}{\nonumber}

\newcommand{\bX}{\bm X}
\newcommand{\bx}{\bm x}

\newcommand{\bZ}{\bm Z}

\newcommand{\bS}{\bm S}
\newcommand{\bs}{\bm s}
\newcommand{\bSigma}{\bm \Sigma}
\newcommand{\bOmega}{\bm \Omega}
\newcommand{\bomega}{\bm \omega}
\newcommand{\balpha}{\bm \alpha}

\newcommand{\btau}{\bm \tau}
\newcommand{\bR}{\bm R}

\newcommand{\bdelta}{\bm \delta}

\DeclareMathOperator*{\argmin}{argmin}

\newcommand{\norm}[1]{\left\lVert#1\right\rVert}

\newcommand{\RR}{\mathbb{R}}

\usepackage[rightbars, color]{changebar}
\newcommand{\rev}[1]{%
  \cbcolor{white}
  \begin{changebar}
    #1
  \end{changebar}%
  }%

\newcommand{\revS}[1]{%
  \cbcolor{white}
  \begin{changebar}
    #1
  \end{changebar}%
  }%

\allowdisplaybreaks[3]
\graphicspath{{../Figures/}}

\begin{document}
\if0\blind
{
  \title{\bf An Expectation Conditional Maximization approach for Gaussian graphical models}
  \author{Zehang Richard Li\thanks{
    We would like to thank Jon Wakefield, Sam Clark, Johannes Lederer, Adrian Dobra, Daniela Witten, and Matt Taddy for helpful discussions and feedback. The authors gratefully acknowledge grants SES-1559778 and DMS-1737673 from the National Science foundation and grant number K01 HD078452 from the National Institute of Child Health and Human Development (NICHD).}\hspace{.2cm}\\
    Department of Biostatistics, Yale School of Public Health\\
    \and
    Tyler H. McCormick \\
    Departments of Statistics \& Sociology, University of Washington}
  \maketitle
} \fi

\if1\blind
{
  \bigskip
  \bigskip
  \bigskip
  \begin{center}
    {\LARGE\bf Title}
\end{center}
  \medskip
} \fi

\bigskip
\begin{abstract}
Bayesian graphical models are a useful tool for understanding dependence relationships among many variables, particularly in situations with external prior information.  In high-dimensional settings, the space of possible graphs becomes enormous, rendering even state-of-the-art Bayesian stochastic search computationally infeasible.  We propose a deterministic alternative to estimate Gaussian and Gaussian copula graphical models using an Expectation Conditional Maximization (ECM) algorithm, extending the EM approach from Bayesian variable selection to graphical model estimation. We show that the ECM approach enables fast posterior exploration under a sequence of mixture priors, and can incorporate multiple sources of information. 
\end{abstract}

\noindent%
{\it Keywords:}  spike-and-slab prior, sparse precision matrix, copula graphical model
\vfill
\newpage
\spacingset{1.45} 

\section{Introduction}
For high dimensional data, graphical models~\citep{lauritzen1996graphical} provide a convenient characterization of the conditional independence structure amongst variables. In settings where the rows in the data matrix $X \in \RR^{n\times p}$ follow an i.i.d multivariate Gaussian distribution, $\mbox{Normal}(\bm 0, \bSigma)$, the zeros in off-diagonal elements of the precision matrix $\bOmega = \bSigma^{-1}$ correspond to pairs of variables that are conditionally independent. Standard maximum likelihood estimators of the sparse precision matrix behave poorly and do not exist when $n < p$, leading to extensive work on algorithms (and their properties) for estimating $\bOmega$~\citep[e.g.,][etc.]{meinshausen2006high,yuan2007model,Friedman2008,RothmanJ.PeterJ.BickelElizavetaLevina2008,Friedman2010,Cai2010,witten2011new,mazumder2012graphical}.


\revS{
In the Bayesian literature, structure learning in high-dimensional Gaussian graphical models has also gained popularity in the past decade. Broadly speaking, two main classes of priors have been studied for inference of the precision matrix in Gaussian graphical models, namely the $G$-Wishart prior, and shrinkage priors. 
The $G$-Wishart prior~\citep{roverato2002hyper} extends the Wishart distribution by restricting its support to the space of positive definite matrices with zeros specified by a graph. It is attractive in Bayesian modeling due to its conjugacy with the Gaussian likelihood. Posterior inference under the $G$-Wishart distribution, though computationally challenging, can be carried out via various algorithms, including shotgun stochastic search~\citep{jones2005experiments}, reversible jump MCMC~\citep{lenkoski2011computational,dobra2012bayesian,Wang2012a}, and birth-death MCMC~\citep{mohammadi2015bayesian}, etc. 
More recently, shrinkage priors for precision matrices have gained much popularity, as they provide Bayesian interpretations to some of the widely used penalized likelihood estimators. As a Bayesian analogy to graphical lasso~\citep{yin2011sparse,witten2011new,mazumder2012graphical}, Bayesian graphical lasso has been proposed in~\citet{wang2012bayesian} and \citet{Peterson2013}. \citet{wang2015scaling} later draws the connection between the Bayesian variable selection~\citep{george1993variable} and Bayesian graphical model estimation, and proposed a new class of spike-and-slab prior for precision and covariance matrices. This class of priors was also later explored in~\citet{Peterson2015} to estimate the dependence structures among regression coefficients, and in~\citet{lukemire2017bayesian} to estimate multiple networks. This type of spike-and-slab prior enables a fast block Gibbs sampler that significantly improves the scalability of the model, but such flexibility is at the cost of prior interpretability since the implied marginal distribution of each elements in the precision matrix is intractable due to the positive definiteness constraint. \citet{wang2015scaling} provides some heuristics and discussions on prior choices, but it is still not clear how to choose the hyperparameters for practical problems or how those choices affect parameter estimation.
}

In this paper, we introduce a new algorithm to estimate sparse precision matrices with spike-and-slab priors~\citep{wang2015scaling} using a deterministic approach, EMGS (EM graph selection), based on the Expectation Conditional Maximization (ECM) algorithm~\citep{meng1993maximum}. We also show that a stochastic variation of the EMGS approach can be extended to copula graphical model estimation. Our work extends the EM approach to variable selection (EMVS)~\citep{rovckova2014emvs} to general graphical model estimation. 

The proposed ECM algorithm is closely connected to frequentist penalized likelihood methods. Similar to the algorithms with concave penalized regularization, such as SCAD~\citep{Fan2009},  the spike-and-slab prior used in our method yields sparse inverse covariance matrix where large values are estimated with less bias (see Figure~\ref{fig:AR1-emgs}). 
\rev{Similar work has been concurrently developed by~\citet{deshpande2017simultaneous} using spike-and-slab lasso prior in the multivariate linear regression models. The proposed approach in this paper differs from~\citet{deshpande2017simultaneous} in two ways: First, we use a mixture of Gaussian distributions instead of the Laplace distributions as the prior on the off-diagonal elements of the precision matrix, which allows us to construct a closed-form conditional maximization step using coordinate descent, rather than relying on additional algorithms solving a graphical lasso problem at each iteration. Second, and more importantly, our work also differs in scope, as we extended the algorithm to non-Gaussian outcomes, the scenarios where informative priors exist, and to incorporate the imputation of missing values.}

\rev{
The rest of the paper is organized as follows: 
In Section~\ref{sec:prior}, we describe the spike-and-slab prior we use for the precision matrix. Section~\ref{sec:ecm} presents the main ECM framework and algorithms for Gaussian graphical model estimation, and Section~\ref{sec:copula} proposes the extension to the copula graphical model and the modified stochastic ECM algorithm. Then in Section~\ref{sec:structure} we explore the incorporation of informative prior knowledge into the model. We discuss briefly about single model selection in Section~\ref{sec:output}. Section~\ref{sec:simulation} examines the performance of our method through numerical simulations. Section~\ref{sec:burke} and \ref{sec:va} further illustrate our model using two examples from scientific settings. Section~\ref{sec:burke} compares our method and alternatives in terms of structure learning and prediction of missing values in a dataset of hourly bike/pedestrian traffic volumes along a busy trail in Seattle. Section~\ref{sec:va} discusses our method in the context of learning latent structures among binary symptoms from a dataset of Verbal Autopsy (VA) surveys, which are used to estimate a likely cause of death in places where most deaths occur outside of medical facilities. 
Finally, in Section~\ref{sec:discuss} we discuss the limitations of the approach and provide some future directions for improvements.
}

\section{Spike-and-slab prior for Gaussian graphical model}\label{sec:prior}
First, we review the \textit{Stochastic Search Structure Learning (SSSL)} prior proposed in~\citet{wang2015scaling} for sparse precision matrices. Consider the standard Gaussian graphical model setting, with observed data $\bX \in \RR^{n\times p}$. Each observation follows a multivariate Gaussian distribution, i.e., $\bx_i \sim \mbox{Normal}(\bm 0, \bOmega^{-1})$, where $\bx_i$ is the $i$-th row of the $\bX$, and $\bOmega$ is the precision matrix. Given hyperparameter $v_0$, $v_1$, and $\pi_{\bdelta}$, the prior on $\bOmega$ is defined as: 
\begin{eqnarray}  
  p(\bOmega | \bdelta)  &=&
  C_{\bdelta}^{-1}\prod_{j<k} \mbox{Normal}(\omega_{jk} | 0, v_{\delta_{jk}}^2)
  \prod_{j}    \mbox{Exp}(\omega_{jj} | \lambda/2) \bm 1_{\Omega \in M^+}  \\ 
p(\bdelta | \pi_{\bdelta}) &\propto& C_{\bdelta}\prod_{j<k} \pi_{\bdelta}^{\delta_{jk}}(1-\pi_{\bdelta})^{1-\delta_{jk}}
  \end{eqnarray}
where $\delta_{jk}$ are latent indicator variables, and $\pi_{\bdelta}$ is the prior sparsity parameter. The $C_{\bdelta}$ term is the normalizing constant that ensures the integration of $p(\bOmega | \bdelta)$ on $M^+$ is one. This formulation places a Gaussian mixture prior on the off-diagonal elements of $\bOmega$, similar to the spike-and-slab prior used in the Bayesian variable selection literature. By setting  $v_1 \gg v_0$, the mixture prior imposes a different strength of shrinkage for elements drawn from the ``slab'' ($v_1$) and ``spike'' ($v_0$) respectively. This representation allows us to shrink elements in $\bOmega$ to $0$ if they are small in scale, while not biasing the large elements significantly. 

\revS{
The spike-and-slab formulation of $\bOmega$ provides an efficient computation strategy via block Gibbs sampling. However, a main limitation is that parameter estimation can be sensitive to the choice of prior parameters. Unlike the variable selection problem in regression, information on the scale of the elements in the precision matrix typically cannot be easily solicited from domain knowledge. As shown in~\citet{wang2015scaling}, there is no analytical relationship between the prior sparsity parameter $\pi_{\bdelta}$ and the induced sparsity from the joint distribution.  This complexity results from the positive definiteness constraint on the precision matrix. Thus even if the sparsity of the precision matrix is known before fitting the model, additional heuristics and explorations are required to properly select the prior $\pi_{\bdelta}$. Similarly, the induced marginal distribution of the elements in $\bOmega$ is intractable as well. The supplementary material contains an simple illustration of such differences. Thus although the fully Gibbs sampler is attractive for high dimensional problems, in practice researchers will usually need to evaluate the model fit under multiple prior choices, adding substantially to the computational burden.
}

\section{Fast deterministic algorithm for graph selection}\label{sec:ecm}
Consider spike-and-slab priors on $\bOmega$ as described in the previous section and let the hyperprior on the sparsity parameter to be $\pi_{\bdelta} \sim \mbox{Beta}(a, b)$, the complete-data posterior distribution can be expressed as
\[
p(\bOmega, \bdelta, \pi_{\bdelta} | \bX) = p(\bX | \bOmega)p(\bOmega | \bdelta, v_0, v_1, \lambda)p(\bdelta|\pi_{\bdelta})p(\pi_{\bdelta}|a, b),
\]
In order to perform posterior sampling in the fully Bayesian fashion, the block Gibbs algorithm in~\citet{wang2015scaling} reduces the problem to iteratively sampling from $(p-1)$-dimensional multivariate Gaussian distributions for each column of $\bOmega$, which can still be computationally expansive for large $p$ or if the sampling needs to be repeated for multiple prior setups. Inspired by the EM approach for variable selection proposed in~\citet{rovckova2014emvs}, we propose a EMGS algorithm to identify the posterior mode of $p(\bOmega, \pi_{\bdelta} | \bX)$ directly without the full stochastic search. We iteratively maximize the following objective function 
\begin{eqnarray}\ok 
Q(\bOmega, \pi_{\bdelta} | \bOmega^{(l)}, \pi_{\bdelta}^{(l)}) 
&=& E_{\bdelta|\bOmega^{(l)}, \pi_{\bdelta}^{(l)}, \bX}(\log p (\bOmega, \bdelta, \pi_{\bdelta} | \bX) | \bOmega^{(l)}, \pi_{\bdelta}^{(l)}, \bX)\\\ok
&=& \mbox{constant} + \frac{n}{2}\log|\bOmega| - \frac{1}{2}tr(\bX^T\bX\bOmega)  \\\ok
&& -\frac{1}{2}\sum_{j < k}\omega_{jk}^2E_{\cdot|\cdot}[\frac{1}{v_0^2(1 - \delta_{jk}) + v_1^2\delta_{jk}}] - \frac{\lambda}{2}\sum_j\omega_{jj} \\\ok
&& + \sum_{j < k}\log(\frac{\pi_{\bdelta}}{1 - \pi_{\bdelta}}E_{\cdot|\cdot}[\delta_{jk}]) + \frac{p(p-1)}{2}\log(1-\pi_{\bdelta})  \\\ok
&&+ (a-1)\log(\pi_{\bdelta}) + (b-1)\log(1-\pi_{\bdelta})
\end{eqnarray}
where $E_{\cdot|\cdot}[\cdot] $ denotes $E_{\bdelta|\bOmega^{(l)}, \pi_{\bdelta}^{(l)}, \bX}[\cdot]$. This objective function can be easily estimated using ECM algorithm, and the algorithm can naturally handle missing values in the E-step. We present the details of the proposed algorithm in the next subsection and then compare the algorithm with the coordinate ascent algorithm for solving graphical lasso problem in Section~\ref{sec:glasso}. 

\subsection{The ECM algorithm}\label{sec:algorithm}
\paragraph{The E-step}
We start by computing the conditional expectations $E_{\bdelta|\bOmega^{(l)}, \pi_{\bdelta}^{(l)}, \bX}[\delta_{jk}]$ and $E_{\bdelta|\bOmega^{(l)}, \pi_{\bdelta}^{(l)}, \bX}[\frac{1}{v_0^2(1 - \delta_{jk}) + v_1^2\delta_{jk}}]$. This proceeds in the similar fashion as the standard EMVS, 
\begin{equation}
E_{\bdelta_{jk}|\bOmega^{(l)}, \pi_{\bdelta}^{(l)}, \bX}[\delta_{jk}] = p^*_{jk} \equiv \frac{a_{jk}}{a_{jk} + b_{jk}},
\end{equation}
where $a_{jk} = p(\omega_{jk} | \delta_{jk} = 1)\pi_{\bdelta}^{(l)}$ and $b_{jk} = p(\omega_{jk} | \delta_{jk} = 0)(1 - \pi_{\bdelta}^{(l)})$, and 
\begin{equation}
E_{\bdelta|\bOmega^{(l)}, \pi_{\bdelta}^{(l)}, \bX}[\frac{1}{v_0^2(1 - \delta_{jk}) + v_1^2\delta_{jk}}] = \frac{1 - p^*_{jk}}{v_0^2} + \frac{p^*_{jk}}{v_1^2} \equiv d^*_{jk}.
\end{equation}
\rev{
\paragraph{Modified E-step with missing data} When missing data exists in the data matrix $\bX$, the E-step can be easily extended to find the expectation of the missing values as well. In that case, the conditional expectations of $\bdelta$ remains unaffected, and we only need to additionally obtain the expectation for the $\bX^T\bX\bOmega$ term as
\begin{equation*}
E_{\bdelta, \bX|\bOmega}(\bX^T\bX\bOmega) =
E_{\bdelta, \bX|\bOmega}(\big(\sum_i^{n} \bx_i\bx_i^T\big)\bOmega) = \big(\sum_i^{n} E_{\bx_{i,m}|\bx_{i,o}, \bOmega}(\bx_i\bx_i^T) \big)\bOmega.
\end{equation*}
where $\bx_{i,o}$ and $\bx_{i,m}$ denote the observed and missing cells in $\bx_i$ respectively. Without loss of generality, if we let $\bx_{i}^T = [\bx_{i,o}^T, \bx_{i,m}^T]$,  we know
 \begin{eqnarray*}
 E_{\bx_{i,m}|\bx_{i,o}, \bOmega}(\bx_{i,m}) &=& 
-\bOmega_{oo}^{-1}\bOmega_{mo}\bx_{i,o}
\\
 E_{\bx_{i,m}|\bx_{i,o}, \bOmega}(\bx_i\bx_i^T) &=&
 E_{\cdot|\cdot}(\bx_i) E_{\cdot|\cdot}(\bx_i)^T + 
 \begin{pmatrix}
    \bm 0_{oo}     & \bm 0_{om} \\
    \bm 0_{mo}       & \bOmega_{mm}^{-1}
\end{pmatrix}
 \end{eqnarray*}
 where $\bOmega_{oo}, \bOmega_{mo}$ and $\bOmega_{mm}$ are the corresponding submatrices of $\bOmega$.
}

\paragraph{The CM-step}
After the E-step is performed, the CM-step performs the maximization of $(\bOmega, \pi_{\bdelta})$ in a coordinate ascent fashion. First, the maximization of $\pi_{\bdelta}$ has the close-form solution
\begin{equation}
\pi_{\bdelta}^{(l+1)} = (a + \sum_{j<k} \delta_{jk} - 1) / (a + b + p(p -1) / 2 - 2).
\end{equation}
The joint maximization of $\bOmega$ has no closed-form solution, but if we denote
 \[
    \bm \Omega = \begin{pmatrix}
                          \bm\Omega_{11} & \bm \omega_{12} \\
                          \bm \omega_{12}^T& \omega_{22}
                         \end{pmatrix}
    \;\;\;\;
    \bX^T\bX = \begin{pmatrix}
                          \bm S_{11} & \bm s_{12} \\
                          \bm s_{12}^T& s_{22}
                         \end{pmatrix}, \]
   \citet{wang2015scaling} showed that the conditional distribution of the last column satisfies
    \[
    \bm\omega_{12} \sim \mbox{Normal}(-\bm{Cs}_{12}, \bm C), \;\;\;\;
    \bm C = ((s_{22} + \lambda)\bm\Omega^{-1} + \mbox{diag}(v_{\bdelta_{12}}))^{-1},
    \]
    where $v_{\bdelta_{12}}$ are the inclusion indicators for $\omega_{12}$ and 
    \[
    \omega_{22} -  \bm\omega_{12}^T\bm\Omega_{11}^{-1}\bm\omega_{12} \sim \mbox{Gamma}(1 + \frac{n}{2}, \frac{\lambda + s_{22}}{2}).
    \]
    This enables us to perform conditional maximization~\citep{meng1993maximum} for the last column holding the rest of $\bOmega$ fixed. That is, starting with $\bOmega^{(l+1)} = \bOmega^{(l)}$, we iteratively permute each column to the last and update it with
    \begin{equation}
    \bm\omega_{12}^{(l+1)} = ((s_{22} + \lambda)(\bOmega_{11}^{(l+1)})^{-1} + \mbox{diag}(d^*_{jk}))^{-1}\bm{s}_{12}
    \end{equation}
  and
    \begin{equation}
    \omega_{22}^{(l+1)} =  (\bm\omega_{12}^{(l+1)})^T(\bOmega_{11}^{(l+1)})^{-1}\bm\omega_{12}^{(l+1)}  + \frac{n}{\lambda + s_{22}}.
    \end{equation}
Finally, be iterating between the E-step and the CM-steps until convergence, we obtain our estimator of the posterior mode $\hat\bOmega$ and $\hat\pi_{\bdelta}$.

\subsection{Connection to the graphical lasso}\label{sec:glasso}
\begin{figure}[!ht]
    \includegraphics[width = .8\textwidth]{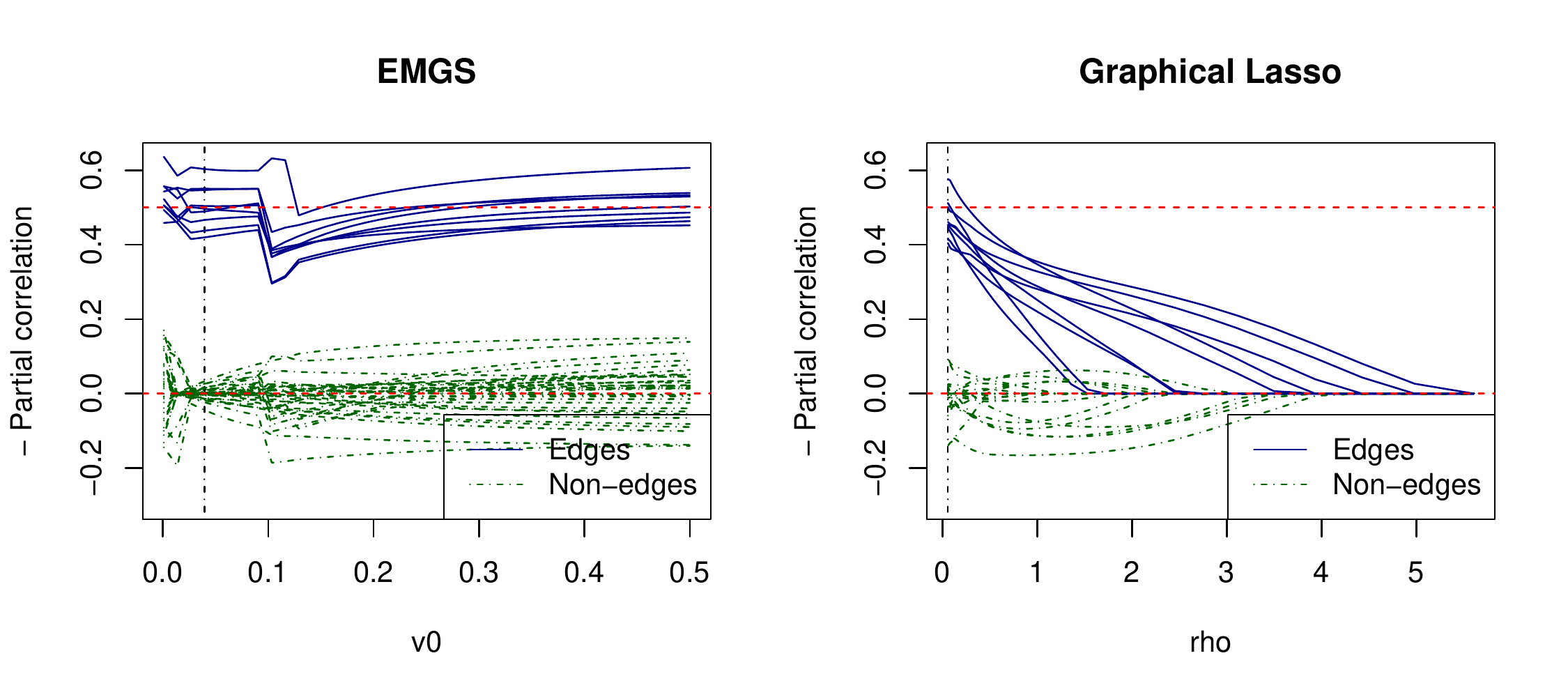}
    \caption{\small Comparing partial correlation path using EMGS and graphical lasso (right), on a $10$-node graph. The red dashed line at $0.5$ is the true value for the non-zero negative partial correlations. The non-zero off-diagonal elements are plotted with blue solid lines. The vertical line indicates the tunning parameter selected with cross-validation.}
    \label{fig:AR1-emgs}
    \end{figure}

This column-wise update resembles the penalized likelihood approach in frequentist settings. In the graphical lasso algorithm~\citep{mazumder2012graphical} for example, the goal is to minimize the $l_1$-penalized negative log-likelihood:
    \[
      f(\bOmega) = -\log|\bOmega| + \mbox{tr}(\bS\bOmega) + \norm{\bOmega}_1,
    \]
    which can be solved via a block coordinate descent that iteratively solves the lasso problem
    \[
      \bomega_{12} = \argmin_{\balpha\in R^{m-1}} \balpha^T\Omega_{11}^{-1}\balpha + \balpha^T\bs_{12} + \lambda \norm{\balpha}_1.
    \]
    The updates at each iteration in the EMGS framework solve the optimization problem for $\bomega_{12}$ under an adaptive ridge penalty
    \[
      \bomega_{12} = \argmin_{\balpha\in R^{m-1}} \balpha^T\bOmega_{11}^{-1}\balpha + \balpha^T\bs_{12} + \sum_{j=1}^{m-1} d^*_{j}\alpha_j^2.
    \]
    The penalty parameters $d^*_{j}$ are the corresponding $d^*_{jk}$ estimated from the E-step and are informed by data. That is, instead of choosing a fixed penalty parameter for all precision matrix elements, the EMGS approach learns the element-wise penalization parameter at each iteration based on the magnitude of the current estimated $\bOmega$ and the hyperpriors placed on $\theta$. Thus, as long as the signal from data is not too weak, the EMGS procedure can estimate large elements in the precision matrix with much lower bias than graphical lasso, as the adaptive penalties associated with large $\bomega_{jk}$ are small. To illustrate the diminished bias, we fit the EMGS algorithm to a simple simulated example, where $n=100$, $p=10$ and $\bOmega$ is constructed by $\omega_{jj} = 1$, and $\omega_{jk} = 0.5$ if $|j-k| = 1$. We fix $v_1 = 100$ and compare the regularization path with various $v_0$ values with graphical lasso, as shown in Figure~\ref{fig:AR1-emgs}. \rev{This simple example illustrates two main advantages of EMGS. First, it identifies the set of non-zero elements quickly and estimates the partial correlations correctly around $0.5$ under all values of $v_0$. The clear separation of the truly non-zero edges regardless of $v_0$ also makes it straightforward to threshold $|\bomega_{jk}|$ to recover the true graph structures. Graphical lasso, on the other hand, shrinks the non-zero partial correlations significantly under large penalties, and thus lead to worse graph selection if the tunning parameter is not properly chosen. Second, In order to select and compare a single model, we also identified the optimal tunning parameter using $5$-fold cross validation for both methods, and it can be seen that the graphical lasso estimator suffers from the weak penalty and contains more noise than using EMGS.} %

\section{ECM algorithm for copula graphical models}\label{sec:copula}
In this section, we extend the framework to non-Gaussian data with Gaussian copulas~\citep{Nelsen1999}. Denote the observed data $\bX \in \RR^{n\times p}$, and each of the $p$ variables could be either continuous, ordinal, or binary. We model each observation as following a Gaussian copula model, i.e., there exists a set of monotonically increasing transformations $f = \{f_1, ..., f_p\}$ such that $\bZ = f(\bX) \sim \mbox{Normal}(\bm 0, \bR)$, where $\bR$ is a correlation matrix. Following the same setup as before, we let $\bR$ be the induced correlation matrix from $\bOmega$ with the spike-and-slab prior defined as before, i.e., 
\[
\bR_{[j,k]} =  \bOmega^{-1}_{[j,k]} / \sqrt{\bOmega^{-1}_{[j,j]}\bOmega^{-1}_{[k,k]}}.
\]
The explicit form of $f$ is typically unknown, thus we impose no restrictions on the class of marginal transformations. Instead, we follow the extended rank likelihood method proposed in~\citet{hoff2007extending}, decomposing the complete data likelihood into
\begin{equation}\label{eqn:rank-likelihood}
  p(\bX | \bR, f) = Pr(\bZ \in \bS | \bR)p(\bX | \bZ \in \bS, \bR, f),
\end{equation} 
where $\bS$ is the support of $\bZ$ induced by the ranking of $\bX$ defined by
\[
  \bS_{ij} = [\max\{z_{i'j'} : x_{i'j'} < x_{ij}\} , \min\{z_{i'j'} : x_{i'j'} > x_{ij}\}].
\]
Since our goal is to recover the structure in $\bOmega$, we can estimate the parameters using only the first part of (\ref{eqn:rank-likelihood}) without estimating the nuisance parameter $f$. 
Moreover, since the latent Gaussian variable $\bZ$ is constructed to be centered at $\bm 0$, the rank likelihood remains unchanged when multiplying columns of $\bX$ by any constant. Thus, inference could be performed without restricting $\bR$ to be an correlation matrix~\citep{hoff2007extending}. In this way, the target function to maximize is the extended rank likelihood function: 
\[
   p(\bOmega, \bdelta, \pi_{\bdelta}, \bZ | \bX) =   p(\bZ \in \bS |\bOmega,\bS)p(\bOmega|\bdelta)p(\bdelta|\pi_{\bdelta}). 
\] 
This is immediately analogous to the EMGS framework with latent Gaussian variable $\bZ$ as additional missing data. 
That is, we maximize the objective function defined as
\begin{eqnarray}\ok 
Q(\bOmega, \pi_{\bdelta}| \bOmega^{(l)}, \pi_{\bdelta}^{(l)}) 
&=& 
E_{\bdelta, \bZ|\bOmega^{(l)}, \pi_{\bdelta}^{(l)}, \bX}(
\log p(\bOmega, \bdelta, \pi_{\bdelta}, \bZ | \bX) 
| \bOmega^{(l)}, \pi_{\bdelta}^{(l)}, \bX)
\\\ok
&=& \mbox{constant} + Q_1 -\frac{1}{2}\sum_{j < k}\omega_{jk}^2E_{\cdot|\cdot}[\frac{1}{v_0^2(1 - \delta_{jk}) + v_1^2\delta_{jk}}] - \frac{\lambda}{2}\sum_i\omega_{ii} \\\ok
&& + \sum_{j < k}\log(\frac{\pi_{\bdelta}}{1 - \pi_{\bdelta}}E_{\cdot|\cdot}[\delta_{jk}]) + \frac{p(p-1)}{2}\log(1-\pi_{\bdelta})  \\\ok
&&+ (a-1)\log(\pi_{\bdelta}) + (b-1)\log(1-\pi_{\bdelta})
\end{eqnarray}

where $E_{\cdot|\cdot}[\cdot] $ denotes $E_{\bdelta, \bZ|\bOmega^{(l)}, \pi_{\bdelta}^{(l)}, \bX}[\cdot]$, and the only term different from the standard EMGS objective function is
\begin{eqnarray}\ok 
Q_1 &=& 
E_{\bZ|\bOmega^{(l)}, \pi_{\bdelta}^{(l)}, \bX}(\log p(\bZ | \bOmega, \bS))
\\\ok &=&
\mbox{constant} +
\frac{n}{2}\log|\bOmega| - \frac{1}{2}E_{\bZ|\bOmega^{(l)},  \bX}[tr(\bZ^T\bZ\bOmega)].
\end{eqnarray}
Exact computation for this expectation is intractable as $\bZ | \bX$ is a Gaussian random matrix where each row is conditionally Gaussian and the within column ranks are fixed by $\bS$. Alternatively, posterior samples of $\bZ$ are easy to obtain from the conditional truncated Gaussian distribution~\citep{hoff2007extending}, so we can adopt stochastic variants of the EM algorithm~\citep{wei1990monte,delyon1999convergence,nielsen2000stochastic,levine2001implementations}. 
We present one such algorithm in the subsequent subsection.

\revS{
\paragraph{The SAE-step for non-Gaussian variables}
Among the many variations of the EM with stochastic approximation, we discuss estimation steps using stochastic approximation EM (SAEM) algorithm~\citep{delyon1999convergence}. SAEM calculates the E-step at each iteration as a weighted average of the current objective function and new stochastic samples using a decreasing sequence of weights for the stochastic averages, in a similar fashion as simulated annealing. In the stochastic E-step, we compute an additional term $Q(\bOmega^{(l)}) = E_{\bZ|\bOmega^{(l)},  \bX}[\bZ^T\bZ]$ as
\[
 Q(\bOmega^{(l)}) = (1 - t_k) Q(\bOmega^{(l)}) + \frac{t_k}{B_k} \sum_{b=1}^{B_k} \bZ_{(b)}^T\bZ_{(b)}
\]
where $t_k$ is an decreasing step-size sequence such that $\sum t_k = \infty$, $\sum t_k^2 < \infty$ ,and $B_k$ is the number of stochastic samples drawn at each iteration. The rank constrained Gaussian variables can be drawn using the same procedure described in~\citet{hoff2007extending}. 

The CM-step then proceeds as before, except that the empirical cross-product matrix $\bS$ is replaced by its expectation $Q(\bOmega^{k})$. For the numerical examples in this paper, we set fixed $B_k$ and $t_k = 1/k$. Other weighting schemes could also be explored and may yield different rates of convergence.
}

\section{Incorporating edge-wise informative priors} \label{sec:structure}
\revS{
The exchangeable beta-binomial prior discussed so far assumes no prior structure on $\bOmega$ and prior sparsity controlled by a single parameter for all off-diagonal elements. 
For many problems in practice, informative priors may exist for pairwise interactions of the variables.  For example, \citet{Peterson2013} infers cellular metabolic networks based on prior information in the form of reference network structures.~\citet{bu2017integrating} improve estimation of brain connectivity network by incorporating the distance between regions of the brain. In problems with small sample sizes, such prior information can help algorithms identify the high probability edges more quickly and provide more interpretable model. 
%
More generally, we can consider a situation where certain groupings exist among variables. For example, when the variables represent log sales of $p$ products on the market, one might expect that the products within the same brand are more likely to be more strongly correlated. If we define a fixed index function $g_j \in \{1, ..., G\}, j \in  \{1, ..., p\}$, where $G$ denotes the total number of groups, we can modify the prior into
\begin{eqnarray}  \ok
  p(\bOmega | \bdelta)  &=&
C_{\bdelta}^{-1}  \prod_{j<k} \mbox{Normal}(\omega_{jk} | 0, \frac{v^2_{\delta_{jk}}}{\tau_{g_jg_k}})
  \prod_{j}    \mbox{Exp}(\omega_{jj} | \lambda/2) \bm 1_{\Omega \in M^+}  \\\ok
p(\bdelta | \pi_{\bdelta}) &\propto& C_{\bdelta}\prod_{j<k} \pi_{\bdelta}^{\delta_{jk}}(1-\pi_{\bdelta})^{1-\delta_{jk}} \\\ok
  p(\btau) &=& \prod_{g<g'}\mbox{Gamma}(a_\tau, b_\tau)
  \end{eqnarray}
The block-wise rescaling parameter $\tau_{g_jg_k}$ of the variance parameter allows us to model within- and between-block elements of $\bOmega$ adaptively with different scales. This is particularly useful in applications where block dependence structures have different strengths. Take the example of sales of products for example. Products within the same brand or category are more likely to be conditional dependent, yet the within group sparsity and the scale of the off-diagonal elements may differ for different brands.} \rev{In the special case where the full edge-level prior probabilities of connection are known, as considered by~\citet{Peterson2013} and \citet{bu2017integrating}, we can also equivalently let $G=P$ and parameterize $p(\btau)$ with the edge-specific priors.} 

\begin{figure}[!ht]
\includegraphics[width = \textwidth]{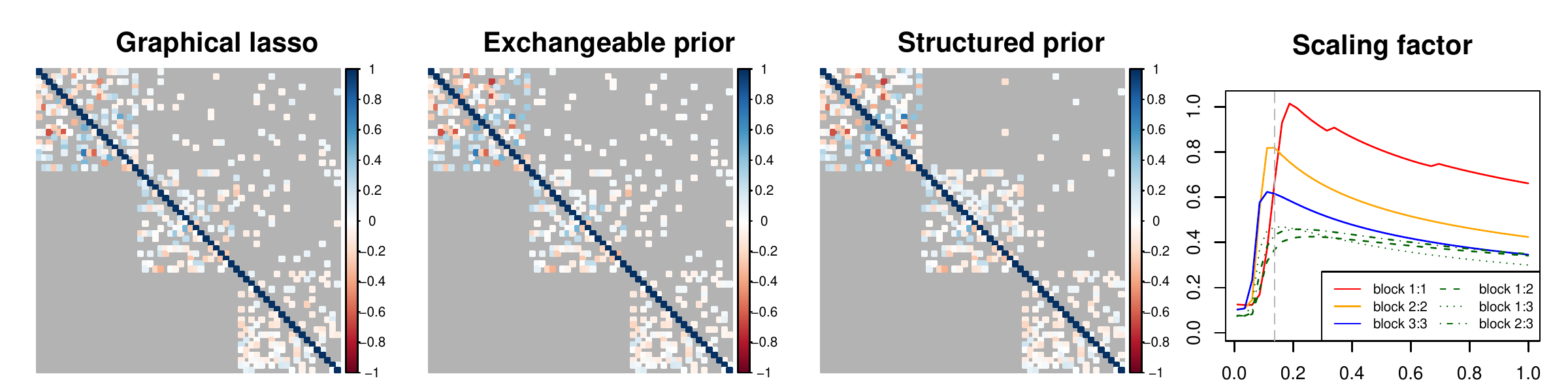}
\caption{Comparing the estimated and true precision matrix using graphical lasso, EMGS with exchangeable prior, and with structured prior for block-wise rescaling. In each plot of the precision matrix comparison, the upper triangle shows the estimated matrix and the lower triangle shows the true precision matrix. All the tunning parameters selected first by cross-validation. The presented edges are thresholded to have the same number of edges compared to the true graph. The forth plot shows the change of $1/\hat\tau_{gg'}$ over different choices of $v_0$. The blocks are labeled $1$ to $3$ from top left to bottom right.}. 
\label{fig:structure}
\end{figure}

\revS{The ECM algorithm discussed above only requires minor modifications to include the additional scale parameter so that the penalties for each block are allowed to vary ~\citep[e.g., ][]{ishwaran2003detecting,wakefield2010bayesian}. The new objective function could be similarly estimated with ECM algorithm by including this additional update in the CM-step:
\begin{equation}
  \tau_{gg'}^{(l+1)} = \frac{a_\tau - 1 + \frac{1}{2}\sum_{j < k} \bm 1_{j, k, g, g'} }{b_\tau + \frac{1}{2}\sum_{j < k} \omega_{jk}^2d_{jk}^*\bm 1_{j, k, g, g'}},
\end{equation}
where $\bm 1_{j, k, g, g'} = 1$ if $g_j=g, g_k=g'$, or $g_j=g', g_k=g$. To illustrate the behavior of this block rescaled prior, we simulate data with 
\rev{$n = 200$, $p = 60$, with the precision matrix to be block diagonal with three equal-sized blocks. We simulate the three block sub-matrices of $\bOmega$ to correspond to random graphs with sparsity $0.4$, as described in Section~\ref{sec:simulation}. Figure~\ref{fig:structure} shows the effect of the structured prior. It can be seen that the estimated $1/\hat\tau_{gg'}$ are much larger where $g = g'$, which leads to weaker shrinkage effects for within cluster cells.}  Accordingly the resulting graph using the structured prior shows fewer false positives for the off-diagonal blocks, and better discovery of the true positives within blocks.  
}

\section{Posterior summary of the ECM output}\label{sec:output}
One of the main computational advantage of the ECM approach over stochastic search is that the posterior mode is fast to obtain. Thus it provides a more efficient alternative to experimenting multiple choices of priors with full MCMC, as discussed before. In practice, we fix $v_1$ to be a large constant and vary the choice of $v_0$ to reflect different levels of shrinkage on the off-diagonal elements of $\bOmega$ that are close to $0$. Intuitively, a larger $v_0$ increases the probability of small parameters being drawn form the spike distribution and thus leads to sparse models. By fitting a sequence of $v_0$, we can create regularization plots, e.g., Figure~\ref{fig:AR1-emgs}, similar to that used in penalized regression literature to visually examine the influence of the prior choices. 
Choosing a single tuning parameter $v_0$ is possible with standard model selection criterion, such as AIC~\citep{akaike1998information}, BIC~\citep{schwarz1978estimating}, RIC~\citep{lysen2009permuted}, StARS~\citep{liu2010stability}, etc., or K-fold cross validation using the average log-likelihood of the validation sets. \rev{In the rest of the paper, we select a single tunning parameter $v_0$ using $5$-fold cross validation. In the case of non-Gaussian data or data with missing values, the likelihood on test data can be evaluated by the average of the expected covariance $\frac{1}{m}\sum_{v_0}E_{\bX_{test}|\bX_{train}, v_0}(\bX_{test}^T\bX_{test})$ under the sequence of $m$ tuning parameters. This term can be easily calculated by plugging in the test data in the E-step of the algorithm.
It is worth noting that since the mixture of Gaussian prior does not lead to exact sparsity, in scenarios where graph structure is of direct interest, we further determining the graph structure by thresholding the off-diagonal elements, $|\omega_{jk}|$, as the posterior inclusion probability $p^*_{jk}$ conditional on $\omega_{jk}$ is a monotone function of $\omega_{jk}$.}




\section{Simulation}\label{sec:simulation}
We follow a similar simulation setup to~\citet{mohammadi2015bayesian} with different graph structures.  We compare the performance of our method with  graphical lasso for Gaussian data and graphical lasso with nonparanormal transformation~\citep{liu2009nonparanormal}, \rev{and the rank-based extension proposed in~\citet{Xue2013} for non-Gaussian data.} We consider the following sparsity patterns in our simulation:
\begin{itemize}
  \item AR(1): A graph with $\sigma_{jk} = 0.7^{|j-k|}$.\vspace{-10pt}
  \item AR(2): A graph with $\omega_{jj} = 1$, $\omega_{j,j-1} = \omega_{j-1, j} = 0.5$, and $\omega_{j,j-2} = \omega_{j-2, j} = 0.25$, and $\omega_{jk} = 0$ otherwise.\vspace{-10pt}
  \item Random: A graph in which the edge set $E$ is randomly generated from independent Bernoulli distributions with probability $0.2$ and the corresponding precision matrix is generated from $\bOmega \sim W_G(3, I_p)$.\vspace{-10pt}
  \item Cluster: A graph in which the number of clusters is $\max\{2, [p/20]\}$. Each cluster has the same structure as a random graph. The corresponding precision matrix is generated from $\bOmega \sim W_G(3, I_p)$.
\end{itemize}
\rev{
We simulate data with sample size $n \in \{100, 200, 500\}$, and of dimension $p \in \{50, 100, 200\}$, using the each types of precision matrices above that are rescaled to have unit variances. We generate both Gaussian and non-Gaussian data for each configuration. For the non-Gaussian case, we perform the marginal transformation of the latent Gaussian variables so that the variables follow a marginal distribution of $\mbox{Poisson}(\theta)$, with $\theta = 10$ or $2$. We simulate graphs with the {\tt R} package {\tt BDgraph}~\citep{mohammadi2015bdgraph}. The graphical lasso estimation are implemented with the {\tt R} package {\tt huge}~\citep{zhao2012huge}.
}

 \rev{
 For each generated graph, we fit our ECM algorithm with a sequence of $40$ increasing $v_0$'s, and fix $v_1=100, \lambda = 1$, and $a=b=1$. We select the final $v_0$ using $5$-fold cross validation.  We also select the tuning parameter for graphical lasso using cross-validation (GL-CV). We then evaluate the bias of EMGS and graphical lasso estimator of precision matrices compared to the truth in terms of the matrix Frobenius norm, $||\hat\bOmega - \bOmega||_F = \sqrt{\sum_{j}\sum_{k}|\hat\omega_{jk} - \omega_{jk}|^2}$. Because of the excess biased induced by a single penalty parameter, cross-validation tend to choose small penalties for graphical lasso, leading to massive false positives in edge discovery. Thus to allow a fair comparison, we compare the area under the ROC curve (AUC) by increasingly thresholding elements in $\hat\bOmega$ obtained by cross-validation for both EMGS and graphical lasso. 
 Besides selecting tuning parameter by cross-validation for graphical lasso and the nonparanormal transformed estimator, we also consider $\hat\bOmega$ selected using two popular model selection criterion: rotaion information criterion (GL-RIC)~\citep{lysen2009permuted}, and stability approach (GL-StARS)~\citep{liu2010stability}. For the copula graphical model, we also compare the rank-based extension proposed in~\citet{Xue2013} of graphical lasso (GL-rank) with the tunning parameter selected with cross validation.
 }

\begin{figure}[!ht]
    \centering
    \includegraphics[page=2, width = \textwidth]{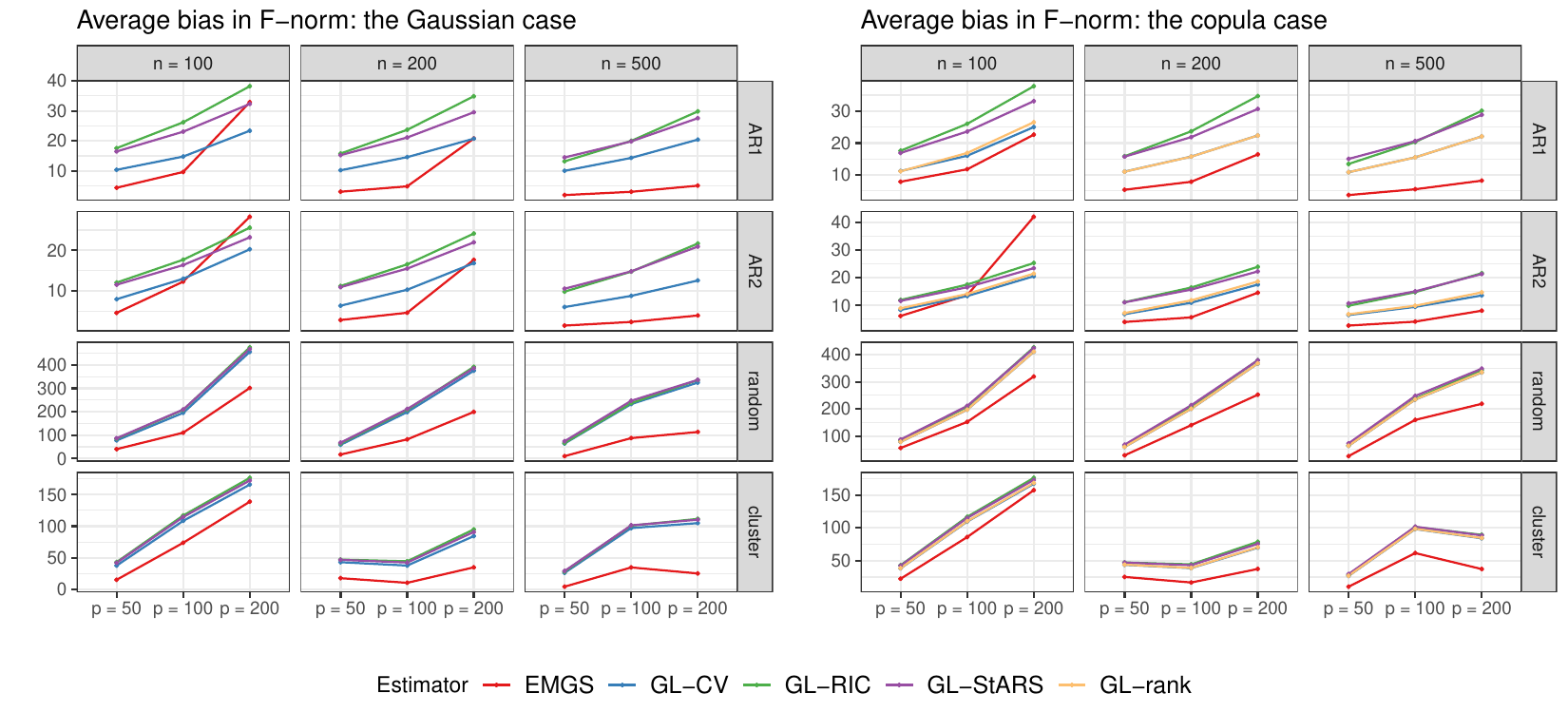}
    \caption{Comparing estimation of the precision matrix for both the Gaussian and Gaussian copula case under different simulation setups. Five estimators are considered: the proposed method (EMGS), gaussian and nonparanormal graphical lasso with penalty selected by cross validataion (GL-CV), RIC (GL-RIC), stability approach (GL-StARS), and rank-based extension of graphical lasso proposed in~\citet{xue2012regularized} selected by cross validation for the copula case (GL-rank).  EMGS shows lower bias in almost all cases.}
    \label{fig:sim-1}
\end{figure}

\begin{figure}[!ht]
    \centering
    \includegraphics[page=2, width = \textwidth]{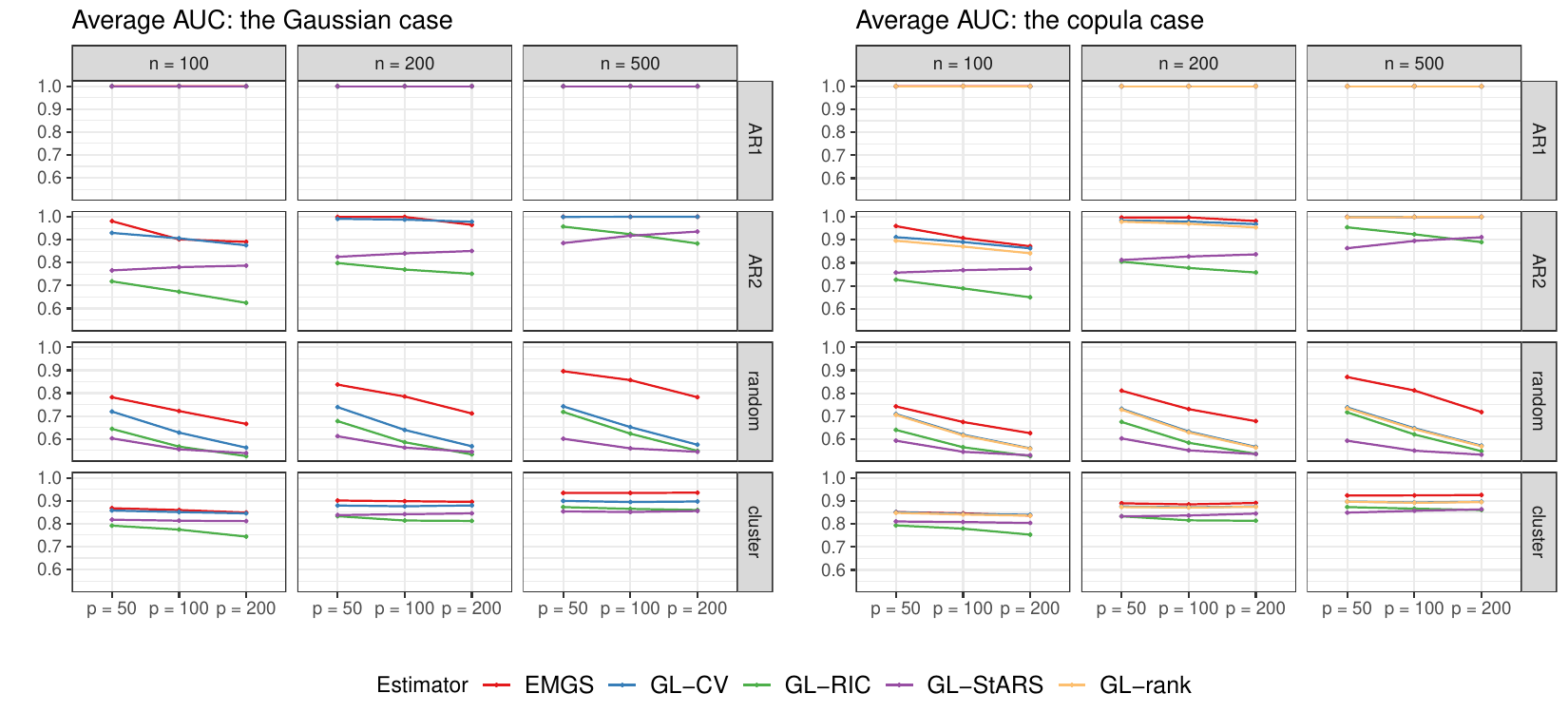}
    \caption{Comparing estimation of the graph structure for both the Gaussian and Gaussian copula case under different simulation setups. EMGS shows higher AUC in almost all cases.}
    \label{fig:sim-2}
\end{figure}

\rev{The simulation results are summarized in Figure \ref{fig:sim-1} and \ref{fig:sim-2}. Less bias in parameter estimation are indicated by smaller $F$-norm values and better graph learning is indicated by larger AUC values. In almost all cases of our simulation study, we observe significantly reduced biases in the estimator from EMGS estimators, as well as better graph selection performance in most cases.
 We also include additional comparisons in the supplementary material that examine the bias in matrix spectral norms, the $F_1$-score for graphical lasso estimators at the selected penalty levels, as well as the $F_1$-score when all estimators are thresholded to have the correct number of edges.
 }

\section{Traffic on the Burke Gilman Trail}
\label{sec:burke}
\rev{In this section we consider graph estimation and prediction for the hourly traffic on the Burke Gilman Trail in Seattle. We use the hourly counts of bikes and pedestrians traveling on the trail through north of NE 70th Street using data from the Seattle Open Data program\footnote{\url{http://www.seattle.gov/tech/initiatives/open-data/}}.} 
\begin{figure}[!ht]
\begin{center}
\includegraphics[width=\textwidth]{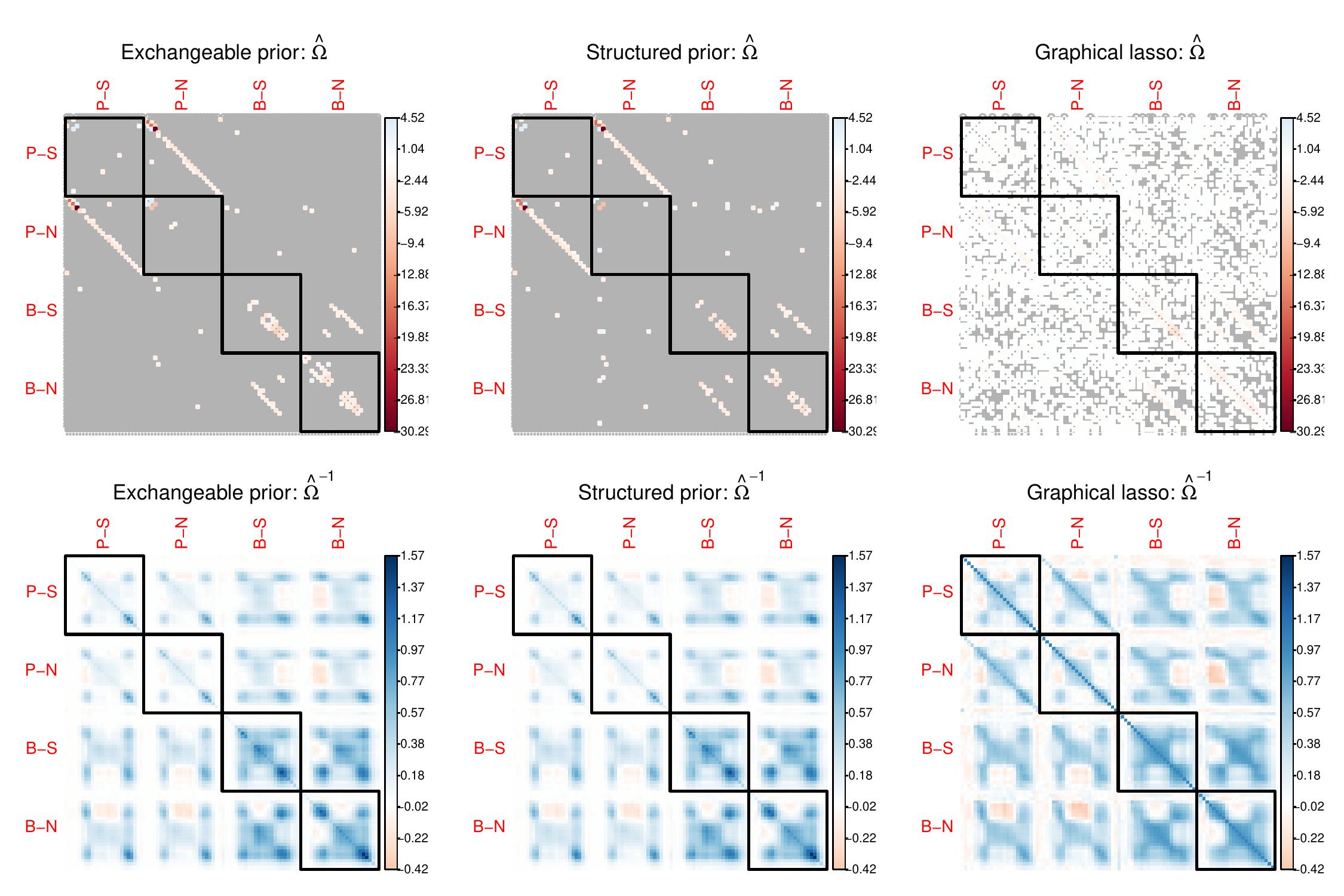}
\end{center}
\caption{\small Comparing the estimated precision matrices from cross validation. The blocks correspond to travel mode and direction pairs. From upper left to lower right: southbound pedestrians, northbound pedestrians, southbound bikes, and northbound bikes. Within each block, the entries correspond to $24$ hourly intervals starting from midnight. \mbox{Top row}: estimated covariance matrix. Edges with less than $0.5$ probability of being from the slab distributions in EMGS output, and exact zeros in graphical lasso output are marked with gray color. \mbox{Bottom row}: estimated precision matrix with highlighted graph selection.}
\label{fig:trail}
\end{figure} 

\rev{The data are captured by sensors that detect both bikes and pedestrians, and their directions of travel. At each hour, the sensors record four counts of travelers: by bike or foot, and towards north or south. We used all the data from 2014 that contain $n = 365$ observations of $24\times4 = 96$ measurements. We first performed a log transformation on the raw counts, and subtracted the hourly average from the log counts. 

We estimated the joint distribution of the $96$ measurements using EMGS with both the beta-binomial prior and the group-wise structured priors, with $4$ groups defined by the mode of travel/direction pairs. Figure~\ref{fig:trail} shows the estimated graphs and the induced covariance matrices. Graphical lasso estimates many edges with small $\omega_{jk}$, while EMGS allows us to pick out large $\omega_{jk}$, especially those that correspond to the edges between the number of pedestrians traveling within the same hour in opposite directions, and the number of bikes traveling in adjacent hours in the same direction during morning and afternoon commute hours. In this analysis, the structured priors lead to a slighly more concentrated set of entires, but both priors lead to similar graph estimation for EMGS. We also compare the performance of predicting missing values using $\hat\Omega$, by randomly removing half of the measurements on half of the days. The missing observations can be imputed by the EMGS algorithm described in Section\ref{sec:ecm}, and similarly we can estimate $\hat\Omega$ by either the empirical covariance matrix or from graphical lasso using only the observed variables. We compare the predictive performance by the Mean Squared Error defined as $MSE = \sum_{i,j}(X_{ij} - \hat{X}_{ij})^2$. }
\begin{table}[!ht]
  \centering
  \begin{tabular}{lllll}
    \toprule
                &\multicolumn{2}{c}{EMGS} & &\\
    \cmidrule(l{2pt}r{2pt}){2-3}
                 & exchangeable& structured  & GLasso & Empirical\\
    \midrule
    Average MSE                   &0.2828&\bf 0.2809& 0.4262&0.4602\\
    Standard deviation of the MSEs&0.0052&\bf 0.0050& 0.0064 & 0.0096\\
    \bottomrule
  \end{tabular}
    \caption{Average and standard deviation of the mean squared errors from 100 cross-validation experiments. }
      \label{tab:mse}
\end{table}

\rev{Intuitively, predictions based on penalized estimators that are over shrunk towards zero is likely to increase bias, while with little penalization, the estimated covariance matrix is more likely to be noisy, as shown in Figure~\ref{fig:trail}. Table~\ref{tab:mse} shows the average MSE and their standard deviations using different estimators over $100$ replications, and it confirms the improved prediction performance from EMGS compared to graphical lasso.    
}

\section{Symptom structure in Verbal Autopsies}
\label{sec:va}
\rev{
In this section, we use EMGS to learn the latent dependence structure among symptoms reported on verbal autopsy (VA) surveys. VA surveys collect information about a deceased person's health history through an interview with caregivers or family members of the decedent. VAs are widely used in countries without full-coverage civil registration and vital statistics systems.  About 2/3 of deaths worldwide occur in such settings~\citep{Horton2007}. VA data consist primarily of binary indicators of symptoms and conditions leading to the death (e.g. Did the decedent have a fever? Was there pain in the lower belly?).}
\begin{figure}[!ht]
\centering
\includegraphics[width=\textwidth]{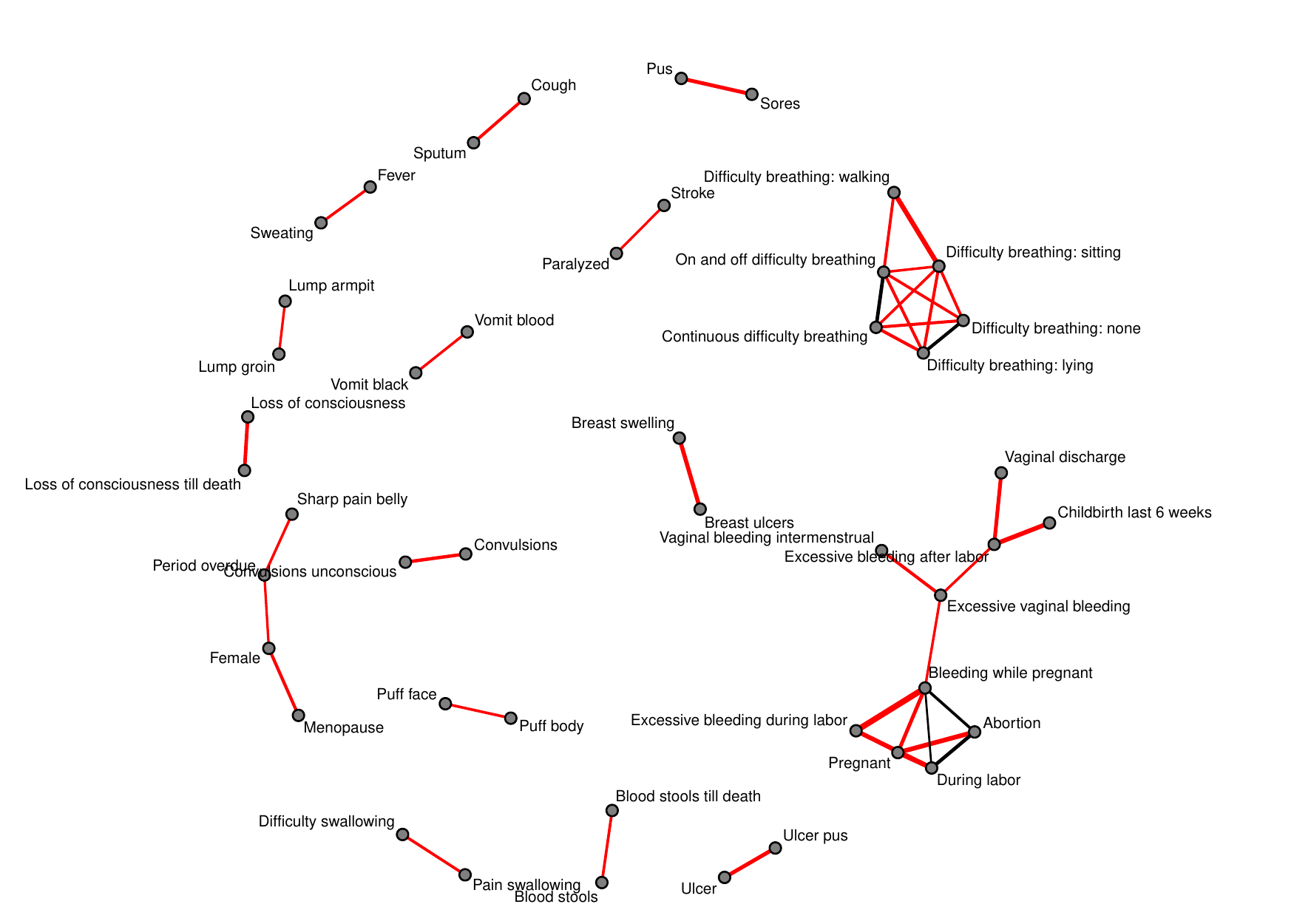}
\vspace{-.5cm}
\caption{Estimated edges between the indicators in the VA dataset. The width of the edges are proportional to the value of $|\omega_{jk}|$. Red edges correspond to negative values of $\omega_{jk}$, or positive partial correlations. Black edges correspond to positive values of $\omega_{jk}$, or negative partial correlations.
 }
\label{fig:va}
\end{figure}

\rev{Several algorithms have been proposed to assign causes of death using such binary input~\citep{byass2012strengthening,serina2015shortened,insilico}, but these algorithms typically assume that the binary indicators are independent. We use data from the Physicians Health Metrics Research Consortium~\citep{murray2011population}. We created $107$ variables from the binary questions in the dataset of $7,841$ adults, and removed the variables with more than $50\%$ of values missing, leaving us with $90$ indicators. There are many missing values even after reducing the number of indicators, so there is only one observation with answers for all $90$ indicators.  This high proportion of missing data makes it difficult to directly apply different types of rank-based estimators for the latent precision matrix that only uses complete observations. Instead, we focus on exploration of the joint distribution of the binary variables under the latent Gaussian framework described in Section~\ref{sec:copula}.  
We first rescale the dataset by the marginal means of the indicators to remove the different levels of prevalence among the symptoms. We then apply the EMGS algorithm to the rescaled dataset with the same hyperpriors used in Section~\ref{sec:simulation}, and select the final $v_0$ using cross validation. The resulting conditional dependence graph with $46$ indicators and $42$ edges is shown in Figure~\ref{fig:va}, where several main symptom pairs (e.g., fever and sweating, stroke and paralysis, etc.) and symptom groups (e.g., indicators related to pregnancy) are discovered, indicating the existence of some symptom clusters that are strongly dependent in the dataset. Further incorporation of the ECM framework into a classification framework could improve accuracy over existing methods for automatic cause-of-death assignment.
}

\section{Discussion}\label{sec:discuss}
\rev{
We propose a deterministic approach for graphical model estimation that builds upon the recently proposed class of spike-and-slab prior for precision matrices. By drawing the connection between the conditional maximization updates under the spike-and-slab prior and the graphical lasso algorithm, we illustrate that EM type algorithm can be used to efficiently obtain posterior modes of the precision matrix under adaptive penalization. It also allows us to build richer class of models that incorporate prior information and extend to copula graphical models.   
The computational speed of the EGMS algorithm allows us to explore multiple prior choices without fitting many time-consuming MCMC chains.
}
However, it also comes at the price of two potential limitations. First, characterization of posterior uncertainty is nontrivial due to the deterministic nature of the algorithm. As in~\citet{rovckova2014emvs}, one may choose to fit a Bayesian model ``locally'' from the posterior mode obtained by the ECM procedure, though this may still be challenging in high-dimensional problems. 
Another limitation is that like the EM algorithm, ECM algorithm also converges only to local modes, thus the precision matrix initialization is critical. In this paper, we used the same initialization as the P-Glasso algorithm described in~\citet{mazumder2012graphical}.  Other heuristics for initialization and warm start may also be explored. Finally, multimodal posteriors are common with spike-and-slab priors. The proposed method could be extended to introduce perturbations in the algorithm, possibly drawing from the variable selection literature~\citep[see, e.g.,][]{rovckova2014emvs,rockova2016particle}.

\rev{
Replication code for the numerical examples in this article is available at \url{https://github.com/richardli/EMGS}.
}

\bibliographystyle{apalike} 
\bibliography{EMbib}

\clearpage
\begin{center}
{\bf \Large Supplementary materials}
\end{center}
\appendix
\section{An illustration of the induced prior in \citet{wang2015scaling}}
\begin{figure}[!ht]
\includegraphics[width=.45\textwidth]{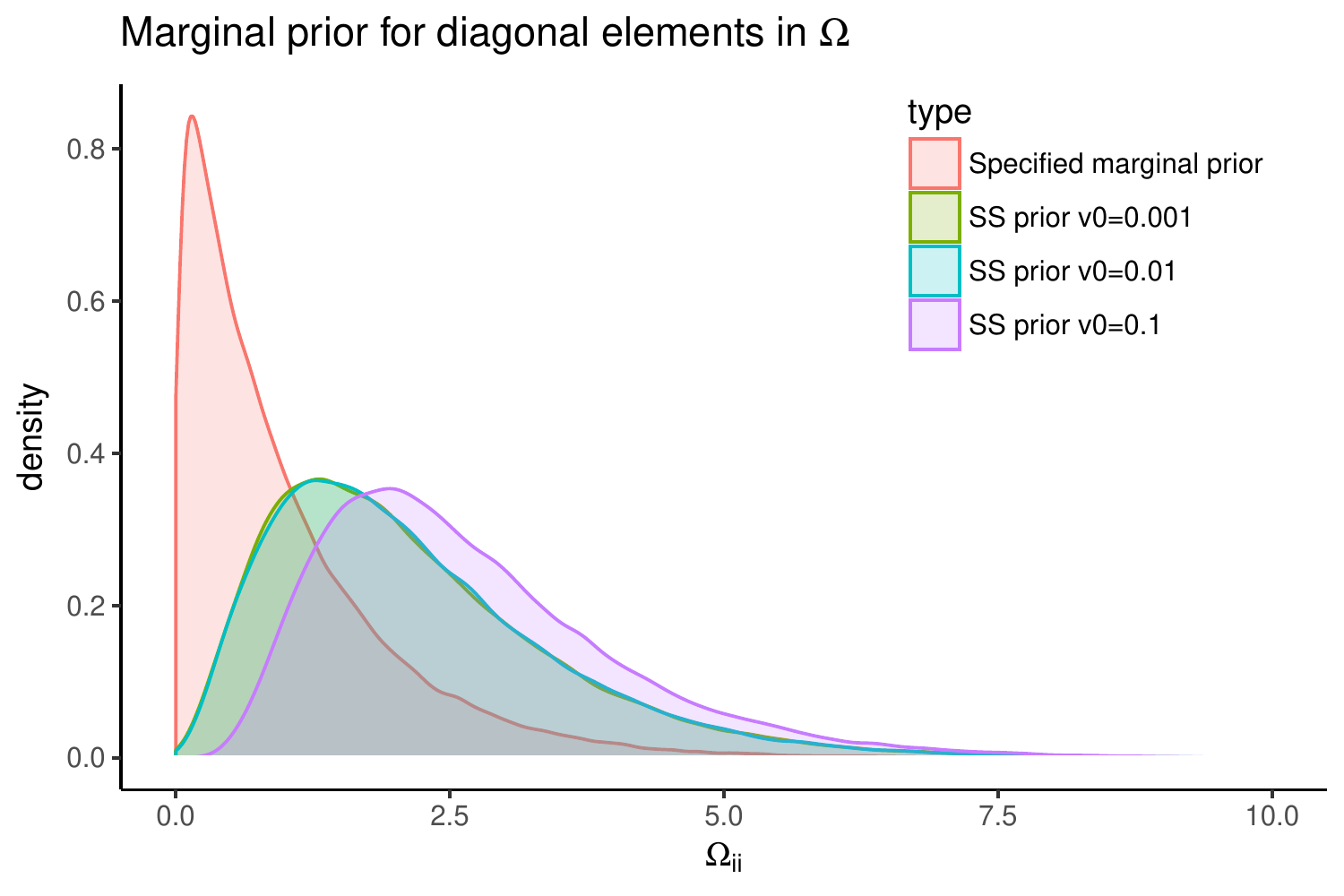}
\includegraphics[width=.45\textwidth]{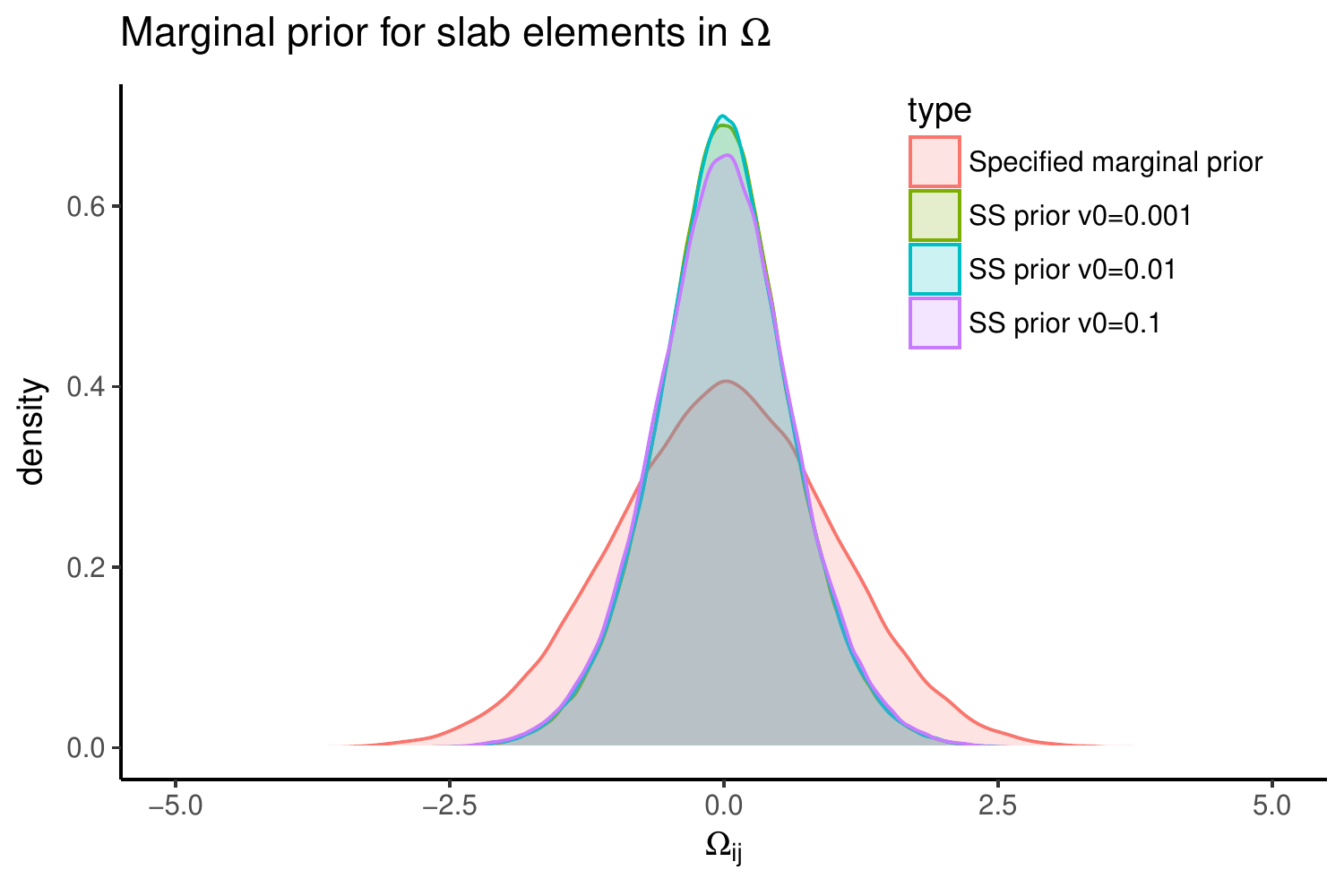}
\caption{Comparison of specified marginal prior distribution and induced marginal prior distributions for $\bOmega$ with $p = 50$, $\lambda = 2$, $v_1 = 1$ and varying $v_0$ values. The underlying graph is fixed to be an AR(2) graph. Left: diagonal elements $\bOmega_{ii}$. Right: Non-zero off-diagonal elements (slab) $\bOmega_{ij}, i \neq j$. The densities are derived from sampling $2,000$ draws using MCMC from the prior distribution after $2,000$ iterations of burn-in.}
\label{fig:prior-1}
\end{figure}

To illustrate the difference between the marginal and induced priors for the precision matrix elements under the formulation of~\citet{wang2015scaling}, Figure~\ref{fig:prior-1} shows several induced marginal distributions for elements of $\bOmega$ when $v_0$ varies and all other parameters are held constant. 

\section{Sampling steps using the rank likelihood}
The SAEM algorithm for the copula graphical model requires sampling the latent Gaussian variables $\bZ |\bOmega, \bX$ in the E-step. The sampling is performed as described in~\citet{hoff2007extending}. The details are as below. For each $j = 1, ..., p$ and $i = 1, ..., n$, given the current values of $\bZ$, we draw new samples of $z_{ij}$ by the following steps:
\begin{enumerate}
        \item find $l = \min(z_{i'j}: x_{i'j} < x_{ij})$, and $u = \min(z_{i'j}: x_{i'j} > x_{ij})$.
        \item compute $m = -\bomega_{j, -j}\bm z_{i, -j}^T/\omega_{jj}$, and  $\sigma^2 = 1/\omega_{jj}$.
        \item draw $q \sim \mbox{Unif}(\Phi(\frac{l-m}{\sigma}), \Phi(\frac{u-m}{\sigma}))$, and set $z_{ij} = m + \sigma\Phi^{-1}(q)$.
\end{enumerate}

\section{The Burke Gilman Trail data}
\begin{figure}[htb]
\includegraphics[width = \textwidth]{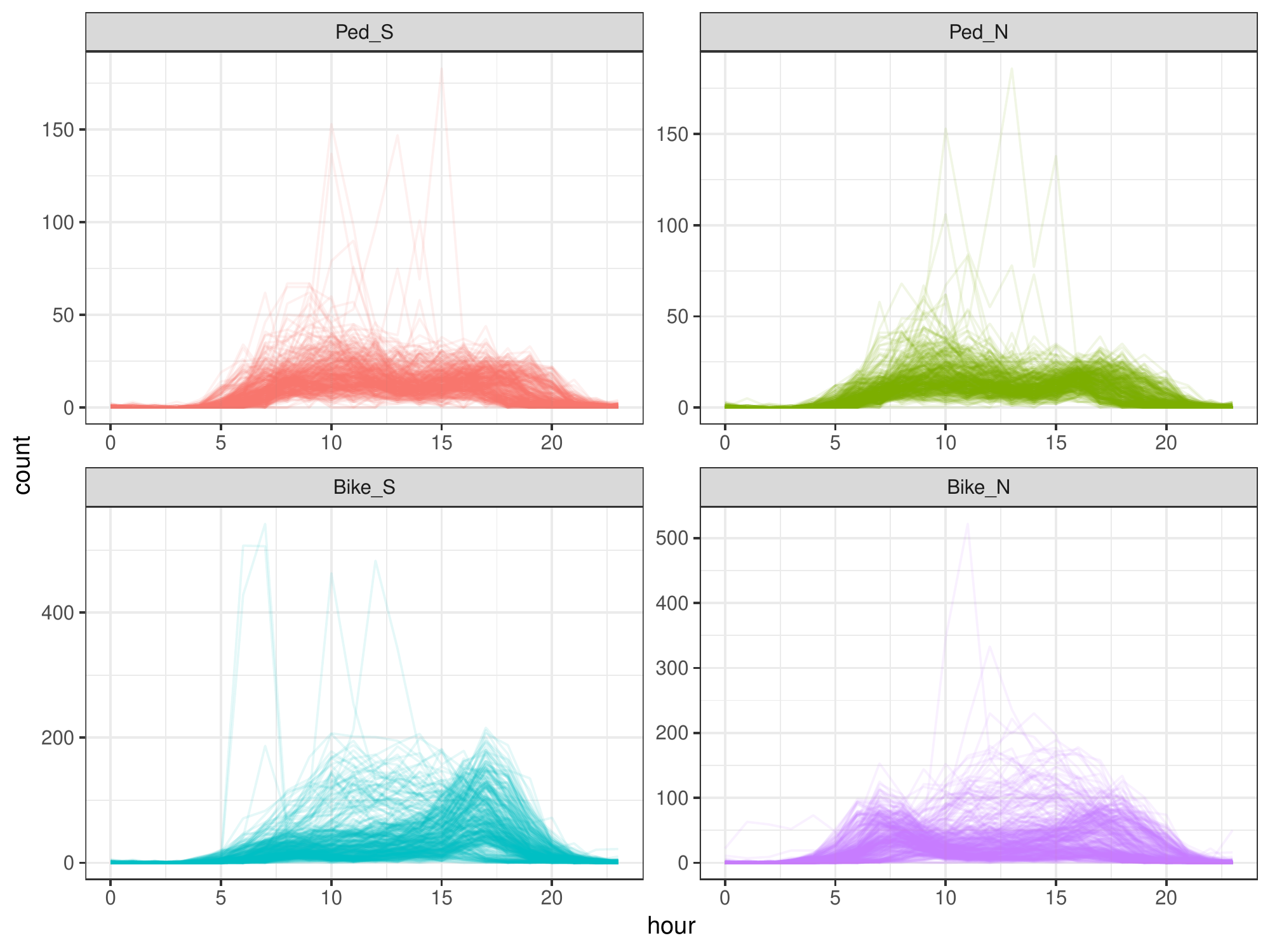}
\caption{Daily volume of travelers captured by sensor at Burke Gilman Trail during 2014 by the four modes of transportation. }
\label{fig:burkeraw}
\end{figure}

A visualization of the daily counts of the four modes of transportation on Burke Gilman Trail is shown in Figure~\ref{fig:burkeraw}. Southbound bike traffic is substantially higher during the afternoon peak hours. This structure has also been learned by the EMGS algorithm under both types of priors.

\section{Additional simulation results}

 Table~\ref{tab:1-1} to \ref{tab:1-4} shows the additional simulation results under each of the four graph structures. Three metrics are summarized in each table: the matrix spectral norms of $||\hat\bOmega - \bOmega||$,  the $F_1$-score by thresholding all estimators to have the same number of edges as in the true graph, and the $F_1$-score at the selected penalty level without thresholding, denoted as $F_1^*$. $F_1$-scores are defined as $F_1 = \frac{2TP}{2TP + FP + FN}$, where $TP, FP$, and $FN$ denote the number of true positive, false positive, and false negative discoveries of edges in the graph. The $F_1$-score can also be written as the harmonic mean of precision and recall. It ranges between $0$ and $1$ where $0$ is the worst case and $1$ is perfect precision and recall. Not surprisingly, in almost all cases, $F_1^*$ is much lower than the $F_1$-score from thresholding to the true sparsity level for any given regularized estimator. EMGS, on the other hand, consistently show similar or higher $F_1$-scores compared to graphical lasso estimators under thresholding. 
\begin{landscape}
\begin{table}[!h]

\caption{\label{tab:}Comparing estimation of the AR1 graph for both the Gaussian and non-Gaussian case. Three different sample sizes,  $n = 100$, $200$, and $500$, are compared in the columns.\label{tab:1-1}}
\centering
\fontsize{7}{9}\selectfont
\begin{tabular}[t]{cccccccccccccccccccc}
\toprule
\multicolumn{1}{c}{ } & \multicolumn{1}{c}{ } & \multicolumn{3}{c}{Gaussian: S-norm} & \multicolumn{3}{c}{Gaussian: $F_1$} & \multicolumn{3}{c}{Gaussian: $F_1^{*}$} & \multicolumn{3}{c}{Copula: S-norm} & \multicolumn{3}{c}{Copula: $F_1$} & \multicolumn{3}{c}{Copula: $F_1^{*}$} \\
\cmidrule(l{2pt}r{2pt}){3-5} \cmidrule(l{2pt}r{2pt}){6-8} \cmidrule(l{2pt}r{2pt}){9-11} \cmidrule(l{2pt}r{2pt}){12-14} \cmidrule(l{2pt}r{2pt}){15-17} \cmidrule(l{2pt}r{2pt}){18-20}
p & Method & 100 & 200 & 500 & 100 & 200 & 500 & 100 & 200 & 500 & 100 & 200 & 500 & 100 & 200 & 500 & 100 & 200 & 500\\
\midrule
50 & EMGS & 1.88 & 1.34 & 0.80 & 1.00 & 1 & 1 &  &  &  & 3.34 & 2.24 & 1.51 & 0.99 & 1 & 1 &  &  & \\
 & GL-CV & 2.95 & 2.79 & 2.67 & 1.00 & 1 & 1 & 0.27 & 0.33 & 0.44 & 3.18 & 3.04 & 2.97 & 1.00 & 1 & 1 & 0.25 & 0.32 & 0.42\\
 & GL-RIC & 4.32 & 3.94 & 3.37 & 1.00 & 1 & 1 & 0.59 & 0.55 & 0.54 & 4.34 & 3.98 & 3.49 & 0.99 & 1 & 1 & 0.58 & 0.54 & 0.52\\
 & GL-StARS & 4.10 & 3.83 & 3.65 & 1.00 & 1 & 1 & 0.52 & 0.54 & 0.56 & 4.20 & 3.96 & 3.82 & 0.99 & 1 & 1 & 0.53 & 0.53 & 0.56\\
 & GL-rank &  &  &  &  &  &  &  &  &  & 3.17 & 3.04 & 2.95 & 1.00 & 1 & 1 & 0.26 & 0.32 & 0.43\\
\hline
100 & EMGS & 2.40 & 1.71 & 1.02 & 1.00 & 1 & 1 &  &  &  & 3.45 & 2.61 & 1.70 & 0.99 & 1 & 1 &  &  & \\
 & GL-CV & 3.01 & 2.87 & 2.71 & 1.00 & 1 & 1 & 0.17 & 0.23 & 0.36 & 3.26 & 3.14 & 3.01 & 1.00 & 1 & 1 & 0.16 & 0.21 & 0.34\\
 & GL-RIC & 4.49 & 4.14 & 3.57 & 1.00 & 1 & 1 & 0.64 & 0.58 & 0.55 & 4.48 & 4.17 & 3.68 & 0.99 & 1 & 1 & 0.62 & 0.56 & 0.54\\
 & GL-StARS & 4.07 & 3.77 & 3.55 & 1.00 & 1 & 1 & 0.45 & 0.48 & 0.55 & 4.17 & 3.92 & 3.73 & 0.99 & 1 & 1 & 0.45 & 0.48 & 0.54\\
 & GL-rank &  &  &  &  &  &  &  &  &  & 3.37 & 3.13 & 3.00 & 1.00 & 1 & 1 & 0.18 & 0.22 & 0.34\\
\hline
200 & EMGS & 4.87 & 3.08 & 1.29 & 1.00 & 1 & 1 &  &  &  & 3.89 & 2.82 & 1.81 & 0.99 & 1 & 1 &  &  & \\
 & GL-CV & 3.30 & 2.93 & 2.76 & 1.00 & 1 & 1 & 0.12 & 0.14 & 0.26 & 3.52 & 3.19 & 3.06 & 0.99 & 1 & 1 & 0.12 & 0.13 & 0.24\\
 & GL-RIC & 4.60 & 4.27 & 3.74 & 0.99 & 1 & 1 & 0.69 & 0.60 & 0.56 & 4.58 & 4.29 & 3.83 & 0.99 & 1 & 1 & 0.66 & 0.58 & 0.55\\
 & GL-StARS & 4.06 & 3.75 & 3.51 & 1.00 & 1 & 1 & 0.34 & 0.40 & 0.53 & 4.18 & 3.92 & 3.71 & 0.99 & 1 & 1 & 0.34 & 0.41 & 0.53\\
 & GL-rank &  &  &  &  &  &  &  &  &  & 3.64 & 3.19 & 3.05 & 0.99 & 1 & 1 & 0.14 & 0.13 & 0.25\\
\bottomrule
\end{tabular}
\end{table}

\begin{table}[!h]

\caption{\label{tab:}Comparing estimation of the AR2 graph for both the Gaussian and non-Gaussian case. Three different sample sizes,  $n = 100$, $200$, and $500$, are compared in the columns.\label{tab:1-2}}
\centering
\fontsize{7}{9}\selectfont
\begin{tabular}[t]{cccccccccccccccccccc}
\toprule
\multicolumn{1}{c}{ } & \multicolumn{1}{c}{ } & \multicolumn{3}{c}{Gaussian: S-norm} & \multicolumn{3}{c}{Gaussian: $F_1$} & \multicolumn{3}{c}{Gaussian: $F_1^{*}$} & \multicolumn{3}{c}{Copula: S-norm} & \multicolumn{3}{c}{Copula: $F_1$} & \multicolumn{3}{c}{Copula: $F_1^{*}$} \\
\cmidrule(l{2pt}r{2pt}){3-5} \cmidrule(l{2pt}r{2pt}){6-8} \cmidrule(l{2pt}r{2pt}){9-11} \cmidrule(l{2pt}r{2pt}){12-14} \cmidrule(l{2pt}r{2pt}){15-17} \cmidrule(l{2pt}r{2pt}){18-20}
p & Method & 100 & 200 & 500 & 100 & 200 & 500 & 100 & 200 & 500 & 100 & 200 & 500 & 100 & 200 & 500 & 100 & 200 & 500\\
\midrule
50 & EMGS & 1.80 & 1.11 & 0.54 & 0.92 & 0.99 & 1.00 &  &  &  & 2.40 & 1.57 & 1.43 & 0.88 & 0.97 & 0.98 &  &  & \\
 & GL-CV & 2.68 & 2.21 & 2.05 & 0.82 & 0.93 & 0.98 & 0.32 & 0.32 & 0.42 & 2.81 & 2.41 & 2.28 & 0.81 & 0.91 & 0.97 & 0.31 & 0.31 & 0.40\\
 & GL-RIC & 3.71 & 3.51 & 3.16 & 0.57 & 0.66 & 0.84 & 0.57 & 0.65 & 0.77 & 3.69 & 3.51 & 3.19 & 0.57 & 0.68 & 0.84 & 0.57 & 0.64 & 0.76\\
 & GL-StARS & 3.59 & 3.45 & 3.35 & 0.59 & 0.75 & 0.81 & 0.58 & 0.65 & 0.72 & 3.62 & 3.49 & 3.40 & 0.58 & 0.71 & 0.80 & 0.57 & 0.64 & 0.70\\
 & GL-rank &  &  &  &  &  &  &  &  &  & 2.99 & 2.56 & 2.41 & 0.79 & 0.90 & 0.96 & 0.34 & 0.32 & 0.41\\
\hline
100 & EMGS & 2.65 & 1.26 & 0.65 & 0.75 & 0.98 & 1.00 &  &  &  & 3.13 & 1.64 & 1.71 & 0.77 & 0.96 & 0.98 &  &  & \\
 & GL-CV & 3.02 & 2.47 & 2.12 & 0.77 & 0.91 & 0.98 & 0.25 & 0.23 & 0.28 & 3.11 & 2.65 & 2.35 & 0.75 & 0.88 & 0.97 & 0.24 & 0.23 & 0.26\\
 & GL-RIC & 3.80 & 3.61 & 3.30 & 0.50 & 0.63 & 0.81 & 0.50 & 0.63 & 0.74 & 3.77 & 3.60 & 3.33 & 0.52 & 0.63 & 0.82 & 0.52 & 0.63 & 0.74\\
 & GL-StARS & 3.59 & 3.44 & 3.32 & 0.70 & 0.78 & 0.81 & 0.52 & 0.62 & 0.74 & 3.62 & 3.48 & 3.38 & 0.69 & 0.77 & 0.81 & 0.51 & 0.61 & 0.72\\
 & GL-rank &  &  &  &  &  &  &  &  &  & 3.26 & 2.86 & 2.48 & 0.73 & 0.86 & 0.96 & 0.27 & 0.26 & 0.27\\
\hline
200 & EMGS & 3.85 & 2.52 & 0.72 & 0.71 & 0.82 & 1.00 &  &  &  & 5.35 & 2.25 & 2.75 & 0.68 & 0.89 & 0.98 &  &  & \\
 & GL-CV & 3.27 & 2.80 & 2.15 & 0.72 & 0.86 & 0.98 & 0.20 & 0.19 & 0.17 & 3.32 & 2.94 & 2.40 & 0.70 & 0.84 & 0.97 & 0.19 & 0.18 & 0.16\\
 & GL-RIC & 3.85 & 3.68 & 3.39 & 0.39 & 0.63 & 0.80 & 0.39 & 0.63 & 0.70 & 3.82 & 3.66 & 3.41 & 0.45 & 0.62 & 0.80 & 0.45 & 0.62 & 0.71\\
 & GL-StARS & 3.59 & 3.44 & 3.30 & 0.68 & 0.77 & 0.82 & 0.42 & 0.55 & 0.74 & 3.63 & 3.49 & 3.38 & 0.66 & 0.76 & 0.81 & 0.41 & 0.55 & 0.72\\
 & GL-rank &  &  &  &  &  &  &  &  &  & 3.44 & 3.10 & 2.60 & 0.68 & 0.81 & 0.95 & 0.22 & 0.22 & 0.18\\
\bottomrule
\end{tabular}
\end{table}

\begin{table}[!h]

\caption{\label{tab:}Comparing estimation of the random graph for both the Gaussian and non-Gaussian case. Three different sample sizes,  $n = 100$, $200$, and $500$, are compared in the columns.\label{tab:1-3}}
\centering
\fontsize{7}{9}\selectfont
\begin{tabular}[t]{cccccccccccccccccccc}
\toprule
\multicolumn{1}{c}{ } & \multicolumn{1}{c}{ } & \multicolumn{3}{c}{Gaussian: S-norm} & \multicolumn{3}{c}{Gaussian: $F_1$} & \multicolumn{3}{c}{Gaussian: $F_1^{*}$} & \multicolumn{3}{c}{Copula: S-norm} & \multicolumn{3}{c}{Copula: $F_1$} & \multicolumn{3}{c}{Copula: $F_1^{*}$} \\
\cmidrule(l{2pt}r{2pt}){3-5} \cmidrule(l{2pt}r{2pt}){6-8} \cmidrule(l{2pt}r{2pt}){9-11} \cmidrule(l{2pt}r{2pt}){12-14} \cmidrule(l{2pt}r{2pt}){15-17} \cmidrule(l{2pt}r{2pt}){18-20}
p & Method & 100 & 200 & 500 & 100 & 200 & 500 & 100 & 200 & 500 & 100 & 200 & 500 & 100 & 200 & 500 & 100 & 200 & 500\\
\midrule
50 & EMGS & 24.01 & 9.40 & 5.58 & 0.72 & 0.79 & 0.86 &  &  &  & 34.11 & 17.33 & 16.52 & 0.67 & 0.76 & 0.83 &  &  & \\
 & GL-CV & 44.90 & 33.36 & 36.61 & 0.66 & 0.69 & 0.70 & 0.46 & 0.48 & 0.48 & 45.32 & 33.74 & 36.11 & 0.65 & 0.68 & 0.69 & 0.44 & 0.47 & 0.47\\
 & GL-RIC & 47.37 & 35.22 & 37.34 & 0.56 & 0.61 & 0.67 & 0.39 & 0.43 & 0.46 & 47.41 & 35.31 & 36.65 & 0.55 & 0.61 & 0.67 & 0.39 & 0.42 & 0.46\\
 & GL-StARS & 48.05 & 36.44 & 39.81 & 0.42 & 0.45 & 0.42 & 0.34 & 0.36 & 0.34 & 48.15 & 36.52 & 39.00 & 0.38 & 0.42 & 0.40 & 0.33 & 0.35 & 0.33\\
 & GL-rank &  &  &  &  &  &  &  &  &  & 45.44 & 33.83 & 36.20 & 0.64 & 0.67 & 0.68 & 0.44 & 0.47 & 0.47\\
\hline
100 & EMGS & 56.73 & 40.86 & 55.68 & 0.65 & 0.73 & 0.82 &  &  &  & 76.57 & 70.38 & 93.69 & 0.59 & 0.66 & 0.76 &  &  & \\
 & GL-CV & 89.36 & 89.54 & 118.14 & 0.55 & 0.56 & 0.58 & 0.38 & 0.39 & 0.41 & 89.83 & 90.01 & 118.91 & 0.53 & 0.56 & 0.58 & 0.37 & 0.39 & 0.40\\
 & GL-RIC & 92.18 & 91.68 & 119.15 & 0.40 & 0.45 & 0.54 & 0.30 & 0.32 & 0.37 & 92.20 & 91.76 & 119.71 & 0.40 & 0.47 & 0.54 & 0.30 & 0.32 & 0.37\\
 & GL-StARS & 92.53 & 92.46 & 121.31 & 0.36 & 0.38 & 0.35 & 0.28 & 0.29 & 0.28 & 92.79 & 92.78 & 121.80 & 0.33 & 0.35 & 0.32 & 0.26 & 0.28 & 0.27\\
 & GL-rank &  &  &  &  &  &  &  &  &  & 89.94 & 90.09 & 119.03 & 0.53 & 0.55 & 0.57 & 0.37 & 0.38 & 0.40\\
\hline
200 & EMGS & 112.14 & 72.03 & 36.88 & 0.57 & 0.63 & 0.72 &  &  &  & 117.70 & 94.47 & 83.09 & 0.52 & 0.59 & 0.64 &  &  & \\
 & GL-CV & 151.84 & 123.53 & 103.07 & 0.28 & 0.29 & 0.30 & 0.28 & 0.29 & 0.30 & 135.26 & 120.00 & 108.31 & 0.28 & 0.29 & 0.30 & 0.28 & 0.29 & 0.30\\
 & GL-RIC & 154.75 & 126.03 & 104.72 & 0.21 & 0.22 & 0.25 & 0.21 & 0.22 & 0.25 & 137.62 & 121.77 & 109.62 & 0.22 & 0.23 & 0.26 & 0.21 & 0.23 & 0.26\\
 & GL-StARS & 153.68 & 125.14 & 104.83 & 0.22 & 0.23 & 0.24 & 0.22 & 0.23 & 0.24 & 137.19 & 121.74 & 110.36 & 0.22 & 0.23 & 0.23 & 0.22 & 0.23 & 0.23\\
 & GL-rank &  &  &  &  &  &  &  &  &  & 135.31 & 120.07 & 108.39 & 0.28 & 0.29 & 0.29 & 0.28 & 0.29 & 0.29\\
\bottomrule
\end{tabular}
\end{table}

\begin{table}[!h]

\caption{\label{tab:}Comparing estimation of the cluster graph for both the Gaussian and non-Gaussian case. Three different sample sizes,  $n = 100$, $200$, and $500$, are compared in the columns.\label{tab:1-4}}
\centering
\fontsize{7}{9}\selectfont
\begin{tabular}[t]{cccccccccccccccccccc}
\toprule
\multicolumn{1}{c}{ } & \multicolumn{1}{c}{ } & \multicolumn{3}{c}{Gaussian: S-norm} & \multicolumn{3}{c}{Gaussian: $F_1$} & \multicolumn{3}{c}{Gaussian: $F_1^{*}$} & \multicolumn{3}{c}{Copula: S-norm} & \multicolumn{3}{c}{Copula: $F_1$} & \multicolumn{3}{c}{Copula: $F_1^{*}$} \\
\cmidrule(l{2pt}r{2pt}){3-5} \cmidrule(l{2pt}r{2pt}){6-8} \cmidrule(l{2pt}r{2pt}){9-11} \cmidrule(l{2pt}r{2pt}){12-14} \cmidrule(l{2pt}r{2pt}){15-17} \cmidrule(l{2pt}r{2pt}){18-20}
p & Method & 100 & 200 & 500 & 100 & 200 & 500 & 100 & 200 & 500 & 100 & 200 & 500 & 100 & 200 & 500 & 100 & 200 & 500\\
\midrule
50 & EMGS & 10.25 & 11.08 & 2.53 & 0.76 & 0.82 & 0.87 &  &  &  & 15.86 & 16.26 & 7.70 & 0.74 & 0.81 & 0.86 &  &  & \\
 & GL-CV & 24.94 & 26.81 & 17.51 & 0.76 & 0.80 & 0.84 & 0.31 & 0.38 & 0.51 & 25.32 & 27.14 & 17.86 & 0.75 & 0.79 & 0.83 & 0.30 & 0.36 & 0.49\\
 & GL-RIC & 27.26 & 28.64 & 18.35 & 0.67 & 0.74 & 0.80 & 0.53 & 0.57 & 0.61 & 27.24 & 28.67 & 18.51 & 0.67 & 0.75 & 0.80 & 0.52 & 0.56 & 0.60\\
 & GL-StARS & 26.89 & 28.47 & 18.82 & 0.73 & 0.75 & 0.77 & 0.51 & 0.55 & 0.61 & 27.03 & 28.63 & 19.02 & 0.72 & 0.74 & 0.77 & 0.51 & 0.55 & 0.61\\
 & GL-rank &  &  &  &  &  &  &  &  &  & 25.48 & 27.24 & 17.99 & 0.75 & 0.79 & 0.83 & 0.31 & 0.36 & 0.50\\
\hline
100 & EMGS & 42.04 & 5.02 & 22.94 & 0.72 & 0.80 & 0.85 &  &  &  & 48.89 & 8.94 & 38.68 & 0.70 & 0.77 & 0.83 &  &  & \\
 & GL-CV & 60.16 & 19.49 & 54.18 & 0.74 & 0.79 & 0.81 & 0.25 & 0.31 & 0.47 & 60.49 & 19.82 & 54.57 & 0.72 & 0.78 & 0.80 & 0.24 & 0.29 & 0.45\\
 & GL-RIC & 62.64 & 21.62 & 55.41 & 0.63 & 0.71 & 0.78 & 0.50 & 0.55 & 0.55 & 62.60 & 21.60 & 55.55 & 0.63 & 0.71 & 0.78 & 0.49 & 0.54 & 0.55\\
 & GL-StARS & 62.00 & 21.06 & 55.53 & 0.71 & 0.76 & 0.76 & 0.44 & 0.52 & 0.54 & 62.13 & 21.23 & 55.69 & 0.70 & 0.75 & 0.76 & 0.44 & 0.52 & 0.54\\
 & GL-rank &  &  &  &  &  &  &  &  &  & 60.69 & 19.95 & 54.68 & 0.72 & 0.78 & 0.80 & 0.25 & 0.29 & 0.45\\
\hline
200 & EMGS & 54.27 & 15.08 & 14.44 & 0.68 & 0.78 & 0.81 &  &  &  & 60.24 & 18.36 & 21.02 & 0.65 & 0.76 & 0.79 &  &  & \\
 & GL-CV & 70.09 & 39.08 & 50.79 & 0.71 & 0.78 & 0.81 & 0.18 & 0.22 & 0.40 & 70.41 & 31.89 & 38.10 & 0.70 & 0.77 & 0.80 & 0.17 & 0.21 & 0.38\\
 & GL-RIC & 72.34 & 41.55 & 52.36 & 0.58 & 0.70 & 0.77 & 0.49 & 0.53 & 0.55 & 72.29 & 33.93 & 39.30 & 0.59 & 0.70 & 0.77 & 0.49 & 0.53 & 0.55\\
 & GL-StARS & 71.54 & 40.68 & 52.04 & 0.69 & 0.74 & 0.75 & 0.37 & 0.44 & 0.53 & 71.69 & 33.28 & 39.09 & 0.68 & 0.74 & 0.77 & 0.36 & 0.44 & 0.54\\
 & GL-rank &  &  &  &  &  &  &  &  &  & 70.72 & 32.06 & 38.24 & 0.69 & 0.77 & 0.80 & 0.19 & 0.22 & 0.38\\
\bottomrule
\end{tabular}
\end{table}
\end{landscape}

\end{document}